\documentclass{article}
\usepackage{spconf,amsmath,graphicx}

\usepackage{enumitem}
\setlist{nosep, leftmargin=14pt}

\usepackage{mwe} 
\usepackage{algorithm}
\usepackage{algpseudocode}
\usepackage{amsmath} 
\usepackage{comment}
\usepackage{graphicx}
\usepackage{xcolor}
\usepackage[colorinlistoftodos]{todonotes}

\title{MTFlow: Time-Conditioned Flow Matching for Microtubule Segmentation in noisy microscopy images}
\name{Sidi Mohamed Sid'El Moctar$^{1}$, Achraf Ait Laydi$^{1,2}$, Yousef El Mourabit$^{2}$, Hélène Bouvrais$^{1}$}
\address{
 $^{1}$ CNRS, Univ. Rennes, Institute of Genetics and Development of Rennes (IGDR), Rennes, France \\
  $^{2}$ TIAD Laboratory, Sciences and Technology Faculty, Sultan Moulay Slimane Univ., Morocco}
\begin{document}
%
\maketitle
\begin{abstract}
Microtubules are cytoskeletal filaments that play essential roles in many cellular processes and are key therapeutic targets in several diseases. Accurate segmentation of microtubule networks is critical for studying their organization and dynamics but remains challenging due to filament curvature, dense crossings, and image noise. We present MTFlow, a novel time-conditioned flow-matching model for microtubule segmentation. Unlike conventional U-Net variants that predict masks in a single pass, MTFlow learns vector fields that iteratively transport noisy masks toward the ground truth, enabling interpretable, trajectory-based refinement. Our architecture combines a U-Net backbone with temporal embeddings, allowing the model to capture the dynamics of uncertainty resolution along filament boundaries. We trained and evaluated MTFlow on synthetic and real microtubule datasets and assessed its generalization capability on public biomedical datasets of curvilinear structures such as retinal blood vessels and nerves. MTFlow achieves competitive segmentation accuracy comparable to state-of-the-art models, offering a powerful and time-efficient tool for filamentous structure analysis with more precise annotations than manual or semi-automatic approaches.
\end{abstract}
\section{Introduction}
\label{sec:intro}

The cytoskeleton is a fundamental component of eukaryotic cells, orchestrating key processes such as intracellular transport, cell division, and morphogenesis \cite{bershadsky2012cytoskeleton}. Among its structural
elements, microtubules play a pivotal role, enabling cells to adapt to external and internal cues. These dynamic tubulin polymers alternate between phases of growth, shrinkage and pause. Precise characterization of microtubule organization and dynamics from microscopy images is crucial for understanding cellular functions and investigating diseases such as cancers or neurodegenerative disorders, where microtubule dysfunction is often implicated \cite{lafanechere2022microtubule, brunden2017altered}.  
However,  automated segmentation of microtubules remains challenging. Microtubules are fine, curvilinear structures, with local curvature, crossings, and sub-resolution widths that confound pixel-level classification. Their networks vary widely in filament length, orientation, and density, and frequently contain overlapping regions that further complicate segmentation. In fluorescence microscopy images, microtubule appearance is also often degraded by photon noise, uneven illumination, and background fluorescence. Similar challenges are encountered when segmenting other curvilinear structures, such as blood vessels or nerves. In addition, the rapid dynamics of microtubules demand short exposure times, resulting in low signal-to-noise images. 

Traditional approaches, such as ridge detectors, steerable filters, and graph-based tracing, have proven useful for curvilinear structure extraction but often rely on hand-crafted parameters that require dataset-specific tuning \cite{kv2023segmentation}. Then, deep learning approaches, particularly convolutional neural networks (CNNs) like U-Net \cite{ronneberger2015u} and its variants, have become the dominant paradigm achieving strong performance by leveraging multi-scale features \cite{kv2023segmentation}.  More recently models incorporating transformers or self-attention mechanisms have further improved global contextual modeling \cite{azad2024advances}.  However, current methods remain limited by high computational demands, reliance on large annotated datasets, and difficulties in capturing fine details, especially in regions with dense or overlapping filaments and in noisy images.

To address these limitations we propose MTFlow, built upon the physics-inspired framework of flow matching \cite{lipman2022flow}. Initially developed for generative modeling, flow matching reformulates prediction as the integration of a learned vector field that incrementally transforms an initial noisy state into the ground-truth segmentation. While flow matching has recently been explored in medical imaging to segment relatively large structures such as polyps or nuclei \cite{bogensperger2025flowsdf, wang2025polypflow}, this work presents, to the best of our knowledge, the first application to tiny and low-contrast structures such as microtubules, nerves and blood vessels in noisy images. Our results showed that the flow-matching dynamic formulation promotes robustness under low signal-to-noise conditions, preserves filament continuity, and improves segmentation efficiency. The main contributions of this work are :

\begin{itemize}
    \item Adapting flow matching dynamics for curvilinear structure segmentation, providing an interpretable alternative to U-Net-based approaches.
    \item Designing a time-conditioned model that learns vector fields to iteratively refine noisy masks into accurate microtubule segmentations.
    \item Demonstrating strong cross-domain generalization by validating our method on public biomedical datasets of curvilinear structures, such as DRIVE, and CORN1, highlighting its potential applications in medical diagnosis.
\end{itemize}

\section{Methodology}
\label{sec:meth}
\subsection{Datasets}
We used the MicSim\_FluoMT synthetic dataset of fluorescently labeled microtubules \cite{bouvrais2025micsim_fluomt}. It contains 1192 images closely resembling real microscopy data with accurately aligned ground-truth masks.
Two variants were used: MicSim\_FluoMT-Simple, with filaments of uniform fluorescence intensity, and MicSim\_FluoMT-Complex, where fluorescence decreases along filaments, posing a more challenging segmentation task. The datasets were split into 953 images for training, 119 for validation, and 120 for testing.

To evaluate MTFlow beyond synthetic images, we used the MicReal\_FluoMT dataset \cite{cueff2025micreal_fluomt}, which includes 49 real fluorescence microscopy images of stained microtubules along with semi-automatically generated masks. Training was conducted using 20 images for training, 9 for validation, and 10 for testing. To test the generalization to other curvilinear structures, we tested MTFlow on two benchmark datasets: DRIVE \cite{staal2004ridge}, which provides 40 images of retinal blood vessels and their corresponding masks, and CORN1 \cite{imedningbo}, which contains 1516 images of corneal nerves and their masks, both of which are manually annotated and publicly available.

\subsection{MTFlow model}

\begin{figure*}
\vspace{-10pt}
\centering
\includegraphics[width=0.999\linewidth, height=5cm]{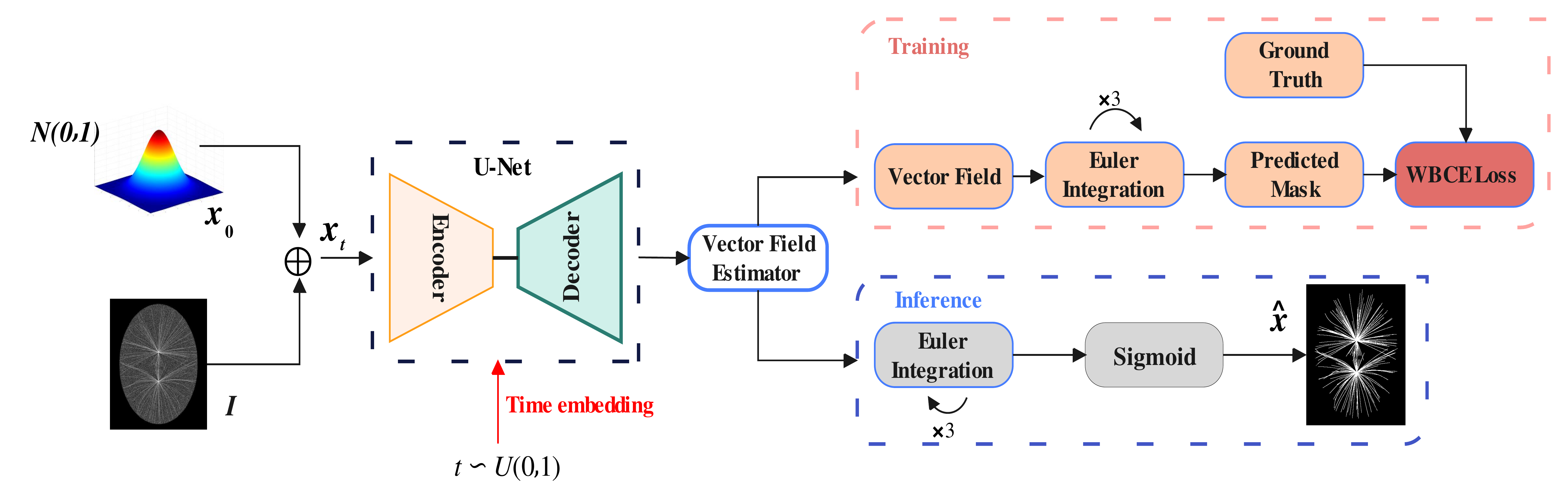}
\caption{\label{fig:mt_flow} \footnotesize An overview of the proposed MTFlow.}
\end{figure*}

We introduce MTFlow, a framework that formulates microtubule segmentation as a dynamic reconstruction problem. Instead of predicting a mask directly, MTFlow learns a time-dependent vector field that iteratively transforms an initial noisy mask toward the ground-truth (Fig. \ref{fig:mt_flow}). MTFlow was built on a U-Net backbone conditioned on the input image, noise, and temporal information. 
Temporal encoding was achieved via sinusoidal embeddings, processed by a small Multi-Layer Perceptron (MLP), and injected into all convolutional blocks to guide the reconstruction at different stages. 
The U-Net consists of four down-sampling and up-sampling blocks, where the number of filters doubles after each down-sampling step (64--128--256--512). 
Each block uses two \(3\times3\) convolutions with GroupNorm (8 groups) and SiLU activations. 
The decoder combines the hierarchical features extracted from the encoder through skip connections to reconstruct spatially precise vector fields that incrementally update the segmentation mask.

Given a microtubule image $I$, we initialized a noisy mask $x_0 \sim \mathcal{N}(0,1)$ and defined a linear interpolation path between $x_0$ and the ground-truth mask $x_1$:
\begin{equation}
x_t = (1-t)x_0 + t x_1, \quad t \in [0,1].
\end{equation}
Conditional flow matching then learned a vector field $v_\theta(x_t, t)$ that aligns with the target displacement:
\begin{equation}
v_\text{target} = x_1 - x_0.
\end{equation}

During inference, the learned dynamics were integrated using an Euler scheme, progressively transforming the noisy initialization into a high-fidelity microtubule mask:
\begin{equation}
x_{n+1} = x_n + \Delta t \, v_\theta(x_n, t_n), \quad n=0,\dots,N-1,
\end{equation}
where $\Delta t$ is the integration step size and $N$ the total number of steps.

The final segmentation was obtained after $N$ steps as:
\begin{equation}
\hat{x} = \sigma \Big( x_0 + \sum_{n=0}^{N-1} \Delta t \, v_\theta(x_n, t_n) \Big)
\end{equation}
where $\sigma$ denotes the sigmoid activation. 

This iterative reconstruction process enabled progressive correction of under-segmented regions and sharpening of microtubule boundaries, provided interpretability, and enhanced robustness to noise and appearence variability.

\subsection{Training Protocol}

We applied standard data augmentation, including random rotations (\(\pm15^{\circ}\)) and horizontal/vertical flips.  
Training was performed using the AdamW optimizer with a batch size of 2, an initial learning rate of \(1\times10^{-4}\), and a weight decay of \(1\times10^{-5}\). 
The learning rate followed a cosine annealing schedule with \(T_{\max}=100\), 
and early stopping with a patience of 30 epochs was applied based on the validation loss to prevent overfitting. 
The computations were performed on a system equipped with NVIDIA RTX A2000 (12 GB) and NVIDIA H100 PCIe (80 GB) GPUs, using CUDA version 12.4.

To handle the class imbalance between microtubules and background pixels, we employed a weighted binary cross-entropy WBCE loss:
{\small
\begin{equation}
\mathcal{L}_\text{WBCE} = \frac{1}{HW} \sum_{i,j} w_1 \, y_{i,j} \log \hat{y}_{i,j} + w_0 \, (1-y_{i,j}) \log (1-\hat{y}_{i,j}),
\end{equation}
}
where $y$ and $\hat{y}$ denote the ground-truth and predicted masks, respectively; $H$ and $W$ are the image height and width; and $w_1 = 1.0$, $w_0 = 0.25$ are the weights for microtubule and background pixels.

\section{Results}
\label{sec:result}
To assess the performance of MTFlow in microtubule segmentation, we compared it with several widely used deep-learning models for biomedical image segmentation, as no existing methods were specifically designed for this task. The selected models included U-Net \cite{ronneberger2015u}, U-Net++ \cite{zhou2018unet}, and ResUNet \cite{zhang2018road}, which have previously been used to segment curvilinear structures, as well as TransUNet \cite{chen2021transunet} and CAR-UNet \cite{guo2021channel}, the latter specifically developed for blood vessel segmentation. All models shared the same backbone as MTFlow, consisting of four down-sampling and up-sampling
blocks. MTFlow, U-Net, U-Net++ and ResUNet were initialized with 64 filters, whereas TransUNet and CAR-UNet started with fewer filters (16 and 32 respectively) to reduce computational load. Segmentation performance was quantified using five metrics, which are reliable under severe class imbalance: Dice, sensitivity, precision, Matthews Correlation Coefficient (MCC), and area under the precision-recall curve (PR-AUC).

\begin{figure}[h]
    \centering
    \includegraphics[width=0.999\linewidth, height=3.2cm]{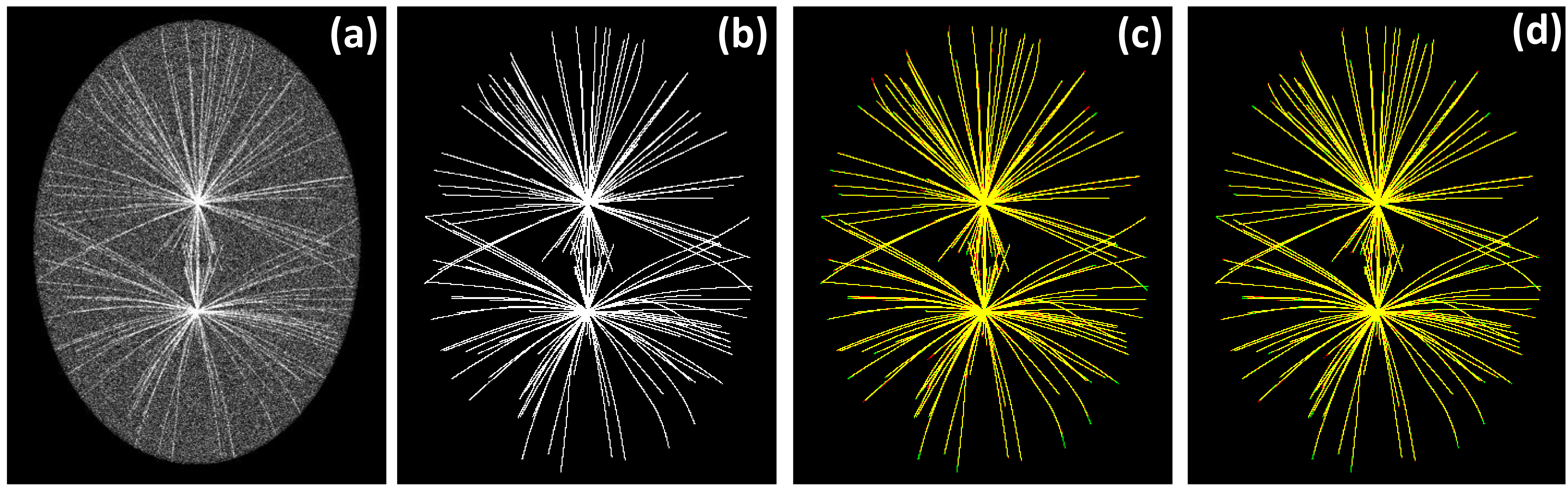}
    \caption{Segmentation on the MicSim\_FluoMT-Simple dataset: (a) Test image, (b) Ground truth, and the composite predictions from (c) MTFlow, and (d) UNet++, in which true positives are in yellow, false positives in red, and false negatives in green.}
    \label{fig:seg_easy}
\end{figure}

\begin{figure}[]
    \centering
    \includegraphics[width=0.999\linewidth, height=3.2cm]{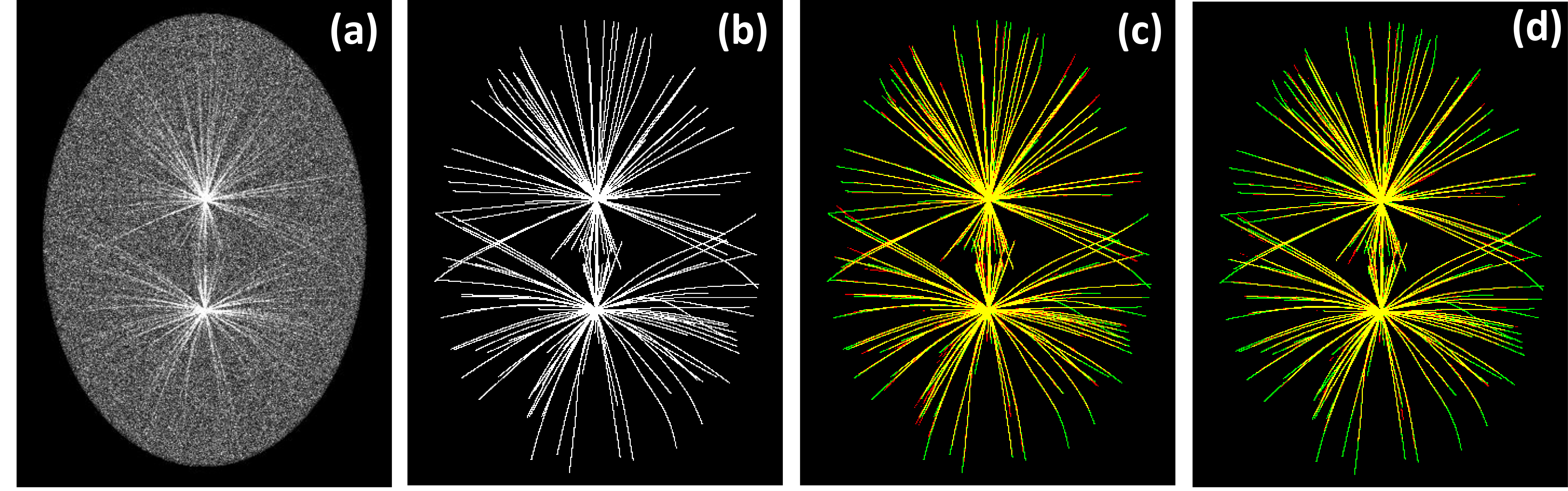}
    \caption{Segmentation on the MicSim\_FluoMT-Complex dataset: (a) Test image, (b) Ground truth, and the composite predictions from (c) MTFlow, and (d) U-Net++, in which true positives are in yellow, false positives in red, and false negatives in green.}
    \label{fig:seg_hard}
\end{figure}

We first evaluated MTFlow on the synthetic microtubule datasets. On the easy dataset, it achieved very accurate segmentations with Dice and MCC, surpassing U-Net, U-Net++, and ResUNet (Fig. \ref{fig:seg_easy}, Table \ref{tab:simple}). CAR-UNet and TransUNet performed notably worse. MTFlow showed higher precision, effectively reducing false positives, at the cost of lower sensitivity, indicating a more conservative and balanced segmentation. 
On the complex dataset, where microtubule extremity intensities are confounded with background fluorescence, MTFlow substantially outperformed all other models (Table \ref{tab:complex}). It achieved the highest Dice, MCC and PR-AUC, along with superior precision. Although other models had higher sensitivity, their precision suffered, indicating a tendency to over-predict. In contrast, MTFlow effectively segmented dense and overlapping filaments without excessive false positives (Fig. \ref{fig:seg_hard}). Overall, MTFlow outperformed other models in segmenting microtubules in noisy images more efficiently. For example, U-Net++, the second-best model, required 4.55 hours of training compared to 3.21 hours for MTFlow.

\begin{table}[h]
\centering
\footnotesize
\caption{Performance on the MicSim\_FluoMT-Simple dataset }
\label{tab:simple}
\begin{tabular}{l|cccccc}
Model & Loss & Dice & Sens. & Prec. & MCC & PR-AUC \\
\hline
U-Net & 0.0052 & 0.9256 & \textbf{0.9812} & 0.8760 & 0.9240 & 0.9886 \\
U-Net++ & 0.0053 & 0.9242 & 0.9807 & 0.8739 & 0.9226 & 0.9882 \\
ResUnet & 0.0055 & 0.9232 & 0.9784 & 0.8735 & 0.9213 & 0.9871 \\
CAR-UNet & 0.0096 & 0.8852 & 0.9587 & 0.8240 & 0.8836 & 0.9688 \\
TransUNet & 0.0112 & 0.8444 & 0.9686 & 0.7484 & 0.8449 & 0.9421 \\
\textbf{MTFlow} & \textbf{0.0051} & \textbf{0.9408} & 0.9431 & \textbf{0.9385} & \textbf{0.9384} & \textbf{0.9887} 
\end{tabular}
\end{table}

\vspace{-10pt}
\begin{table}[H]
\centering
\footnotesize
\caption{Performance on the MicSim\_FluoMT-Complex dataset }
\label{tab:complex}
\begin{tabular}{l|cccccc}
Model & Loss & Dice & Sens & Prec & MCC & PR-AUC \\
\hline
U-Net & 0.0202 & 0.7681 & 0.8953 & 0.6725 & 0.7658 & 0.8939 \\
U-Net++ & 0.0195 & 0.7788 & 0.8942 & 0.6894 & 0.7755 & 0.9007 \\
ResUnet & 0.0194 & 0.7672 & \textbf{0.9067} & 0.6649 & 0.7663 & 0.8974 \\
CAR-UNet & 0.0472 & 0.6278 & 0.7532 & 0.5621 & 0.6261 & 0.7281 \\
TransUNet & 0.0309 & 0.6389 & 0.8528 & 0.5107 & 0.6432 & 0.7258 \\
\textbf{MTFlow} & \textbf{0.0170} & \textbf{0.8228} & 0.7867 & \textbf{0.8624} & \textbf{0.8169} & \textbf{0.9147} 
\end{tabular}
\end{table}

\vspace{-3pt}

\begin{figure}[H]
    \centering
    \includegraphics[width=0.9999999\linewidth, height=3cm]{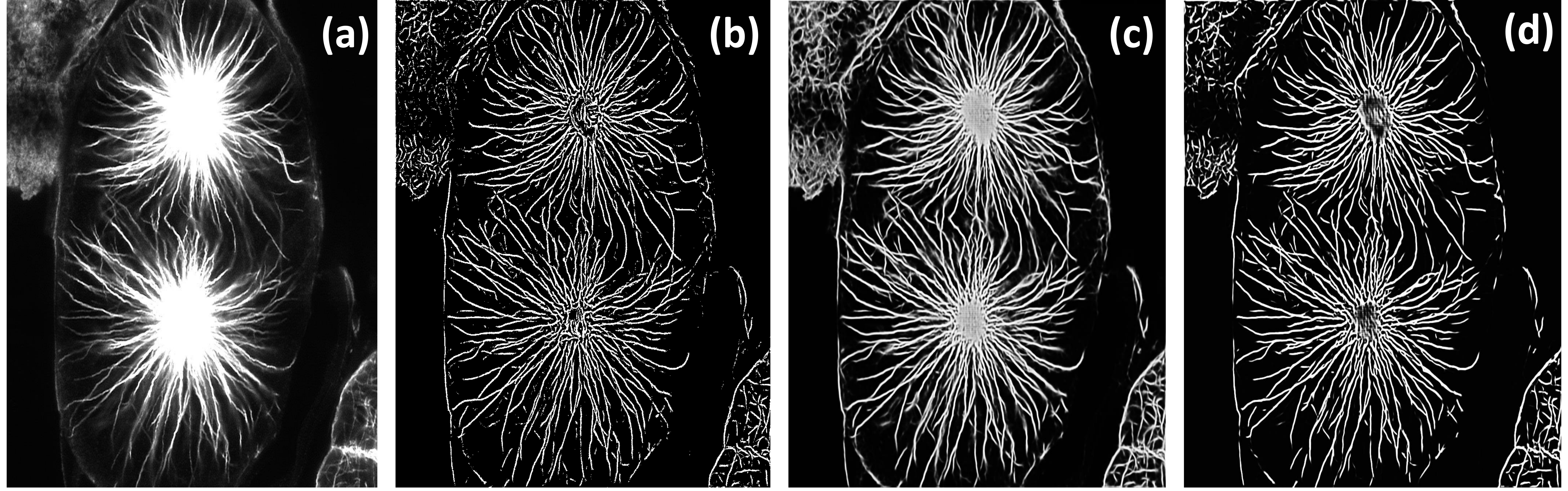}
    \caption{Segmentation on the MicReal\_FluoMT dataset: (a) Test image, (b) Ground truth, and the predictions from (c) MTFlow, and (d) U-Net++.}
    \label{fig:seg_real}
\end{figure}

We then challenged MTFlow on real microscopy images using the MicReal\_FluoMT dataset, which poses challenges due to fluorescence intensity variability among microtubules. MTFlow produced accurate and visually clean segmentations, demonstrating its robustness (Fig. \ref{fig:seg_real}). Notably, the predicted segmentations appeared less noisy than the annotations. These observations were consistent with the quantitative results (Table \ref{tab:micreal}), where MTFlow achieved performance comparable to U-Net++. Importantly, MTFlow achieved a markedly superior balance between precision and recall, as reflected by its higher PR-AUC, enabling it to capture true microtubule structures while effectively limiting erroneous detections (Fig. \ref{fig:seg_real}). It is worth noting that some MTFlow predcitions (e.g. at the spindle poles) were not well annotated and therefore counted as false positives, slightly underestimatin MTFlow's precision. Overall, MTFlow generalized effectively to real images, accurately segmenting microtubules despite annotation noise and intensity variability.

\vspace{-5pt}
\begin{table}[H]
\centering
\footnotesize
\caption{Performance on the MicReal\_FluoMT dataset}
\label{tab:micreal}
\begin{tabular}{l|cccccc}
Model & Loss & Dice & Sens & Prec & MCC & PR-AUC \\
\hline
U-Net & 0.0884 & 0.6617 & 0.6434 & 0.6810 & 0.6297 & 0.7165 \\
U-Net++ & 0.0996 & \textbf{0.7038} & 0.6886 & \textbf{0.7197} & \textbf{0.6755} & 0.7444 \\
\textbf{MTFlow} & \textbf{0.0685} & 0.6968 & \textbf{0.6916} & 0.7021 & 0.6673 & \textbf{0.7736} \\

\end{tabular}
\vspace{-3pt}
\end{table}

\begin{figure}[htbp]
    \centering
    \includegraphics[width=1\linewidth, height=4.2cm]{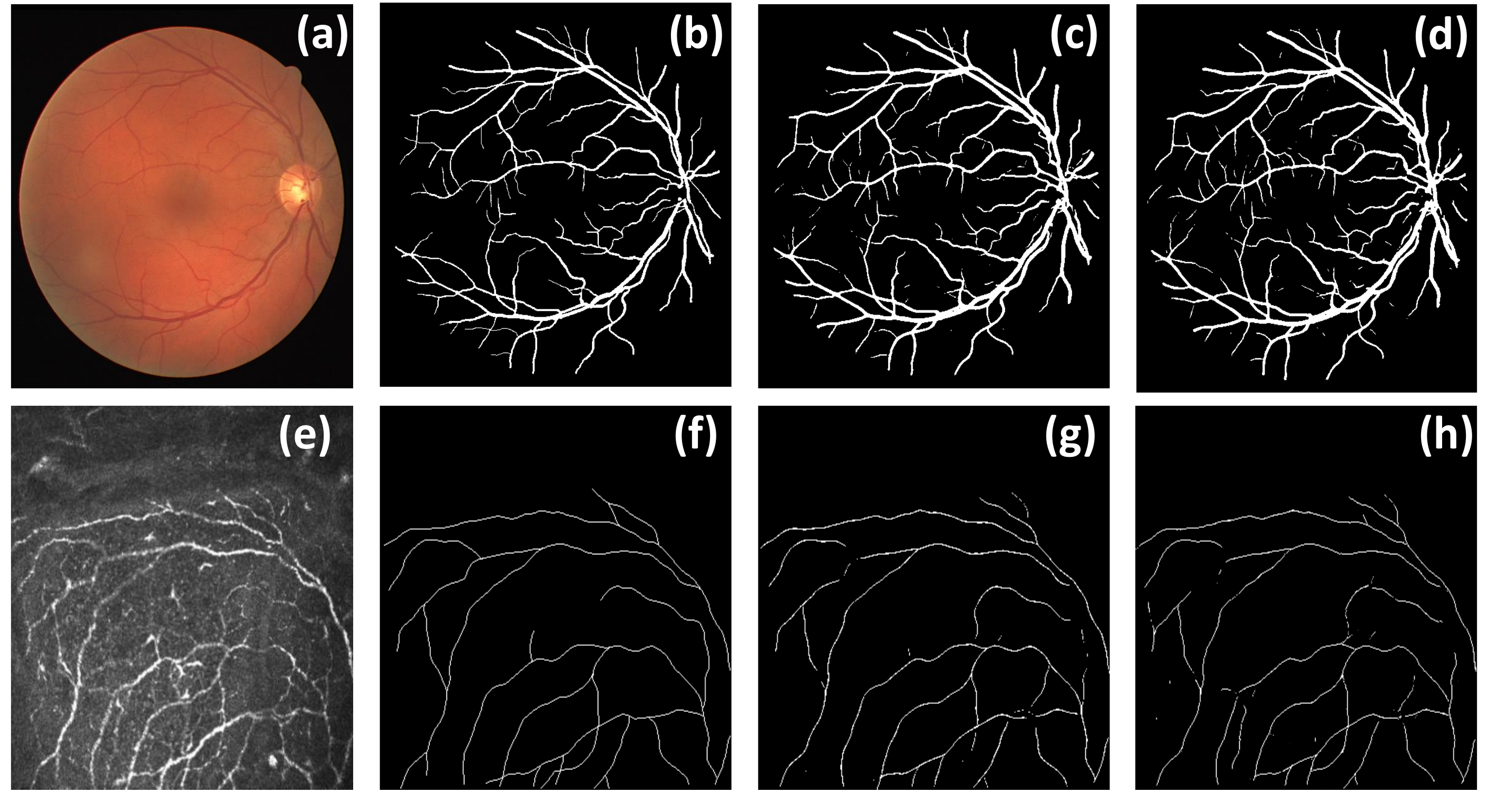}
    \caption{Segmentation on the DRIVE (a-d) and the CORN1(e-h) datasets: (a,e) Test image, (b,f) Ground truth, and the predictions from (c,g) MTFlow, and (d,h) U-Net++.}
    \label{fig:seg_drive_corn1}
    \vspace{-8pt}
\end{figure}

We last wondered whether MTFlow would succeed in segmenting different types of curvilinear structures, in particular blood vessels and nerves.
On the DRIVE retinal vessel dataset, MTFlow demonstrated superior performance compared to U-Net and U-Net++. It achieved the highest Dice, MCC, and PR-AUC, indicating a better-balanced trade-off between false positives and false negatives (Table \ref{tab:Drive_CORN1} and Fig. \ref{fig:seg_drive_corn1}a-b). Indeed, although MTFlow's precision was slightly lower than that of U-Net, reflecting a minor increase in false positives, it achieved higher sensitivity, suggesting a better vessel coverage than U-Net. U-Net++ led to the best sensitivity but at the expense of the lowest precision. On the CORN1 corneal nerve dataset, MTFlow effectively segmented nerves (Fig. \ref{fig:seg_drive_corn1}e-h) and outperfomed U-Net and U-Net++ across all metrics (Table \ref{tab:Drive_CORN1}). 
By successfully segmenting both corneal nerves and retinal vessels, MTFlow demonstrated strong generalization beyond microtubules.

\begin{table}[htbp]
\vspace{-5pt}
\centering
\footnotesize
\caption{Performance comparison on the DRIVE and the CORN1 datasets}
\label{tab:Drive_CORN1}
\begin{tabular}{l|c|ccccc}
\textbf{Dataset} & \textbf{Model} & \textbf{Dice} & \textbf{Sens} & \textbf{Prec} & \textbf{MCC} & \textbf{PR-AUC} \\
\hline
 & U-Net & 0.8021 & 0.8247 & \textbf{0.7806} & 0.7838 & 0.8888 \\
\textbf{DRIVE} & U-Net++ & 0.8076 & \textbf{0.8620} & 0.7597 & 0.7909 & 0.8989 \\
 & \textbf{MTFlow} & \textbf{0.8106} & 0.8493 & 0.7752 & \textbf{0.7934} & \textbf{0.9003} \\
\hline
 & U-Net & 0.7699 & 0.7539 & 0.7866 & 0.7666 & 0.7635 \\
\textbf{CORN1} & U-Net++ & 0.7722 & 0.7580 & 0.7870 & 0.7689 & 0.7714 \\
 & \textbf{MTFlow} & \textbf{0.7747} & \textbf{0.7610} & \textbf{0.7890} & \textbf{0.7715} & \textbf{0.7846} \\
\end{tabular}
\vspace{-8pt}
\end{table}

\vspace{-5pt}

\section{Conclusion}
\label{sec:concl}

We present MTFlow, a time-conditioned flow matching framework for segmenting curvilinear structures in noisy microscopy or biomedical images. By iteratively refining noisy masks through learned vector fields, MTFlow provides accurate and balanced reconstructions, effectively handling thin, low-contrast, and overlapping filaments. Evaluations on synthetic and real microtubule datasets, as well as retinal vessel and corneal nerve images, demonstrate that MTFlow consistently delivers precise and robust segmentations, surpassing U-Net and its variants while maintaining generalization across diverse datasets. These results establish MTFlow as a reliable and interpretable tool for the quantitative analysis of curvilinear structures in both research and clinical contexts, facilitating cellular studies through cytoskeletal filament segmentation and aiding diagnosis based on nerve and vessel morphology. 
Future work will explore architectural refinements, integration of attention mechanisms, uncertainty-aware training to sharpen filament boundaries, and self-supervised learning to reduce reliance on densely annotated data.

\section{Acknowledgment}
This work was supported by the Agence Nationale de la Recherche (ANR-22-CE45-001601) and by Campus France/ CNRST (PHC Toubkal 2024, n° 49945RE).

\bibliographystyle{IEEEbib}
\bibliography{refs}

\end{document}